\documentclass[letterpaper]{article} 
\usepackage{aaai24}  
\usepackage{times}  
\usepackage{helvet}  
\usepackage{courier}  
\usepackage[hyphens]{url}  
\usepackage{graphicx} 
\usepackage{natbib}  
\usepackage{caption} 
\usepackage{algorithm}
\usepackage{algorithmic}
\usepackage{newfloat}
\usepackage{listings}
\usepackage{amsmath}
\usepackage{amssymb}
\usepackage{multirow}
\usepackage{xcolor}

\usepackage{newfloat}
\usepackage{listings}
\DeclareCaptionStyle{ruled}{labelfont=normalfont,labelsep=colon,strut=off} 
\floatstyle{ruled}
\newfloat{listing}{tb}{lst}{}
\floatname{listing}{Listing}
\title{Enhancing RAW-to-sRGB with Decoupled Style Structure in Fourier Domain}
\author {
    Xuanhua He\textsuperscript{\rm 1,\rm 2}\thanks{Co-first authors contributed equally. This work was done when Xuanhua He was an intern at Horizon Robotics. $^\dagger$ Corresponding author.},
    Tao Hu\textsuperscript{\rm 1,\rm 2}\footnotemark[1],
    Guoli Wang\textsuperscript{\rm 3},
    Zejin Wang\textsuperscript{\rm 3},
    Run Wang\textsuperscript{\rm 3},
    Qian Zhang\textsuperscript{\rm 3},
    Keyu Yan\textsuperscript{\rm 1,\rm 2},\\
    Ziyi Chen\textsuperscript{\rm 4},
    Rui Li\textsuperscript{\rm 1},
    Chenjun Xie\textsuperscript{\rm 1},
    Jie Zhang\textsuperscript{\rm 1}$^{\dagger}$,
    Man Zhou\textsuperscript{\rm 5}$^{\dagger}$
}
\affiliations {
    \textsuperscript{\rm 1}Hefei Institutes of Physical Science, Chinese Academy of Sciences
    \textsuperscript{\rm 2}University of Science and Technology of China\\
    \textsuperscript{\rm 3}Horizon Robotics
    \textsuperscript{\rm 4}Tencent Technology
    \textsuperscript{\rm 5}Nanyang Technological University\\
    {\{hexuanhua,ht\_simon,keyu\}@mail.ustc.edu.cn, \{cjxie,zhangjie\}@iim.ac.cn,manzhountu@gmail.com}\\
}

\usepackage{bibentry}

\begin{document}

\maketitle

\begin{abstract}
RAW to sRGB mapping, which aims to convert RAW images from smartphones into RGB form equivalent to that of Digital Single-Lens Reflex (DSLR) cameras, has become an important area of research. However, current methods often ignore the difference between cell phone RAW images and DSLR camera RGB images, a difference that goes beyond the color matrix and extends to spatial structure due to resolution variations. Recent methods directly rebuild color mapping and spatial structure via shared deep representation, limiting optimal performance. Inspired by Image Signal Processing (ISP) pipeline, which distinguishes image restoration and enhancement, we present a novel Neural ISP framework, named FourierISP. This approach breaks the image down into style and structure within the frequency domain, allowing for independent optimization. FourierISP is comprised of three subnetworks: Phase Enhance Subnet for structural refinement, Amplitude Refine Subnet for color learning, and Color Adaptation Subnet for blending them in a smooth manner. This approach sharpens both color and structure, and extensive evaluations across varied datasets confirm that our approach realizes state-of-the-art results. Code will be available at ~\url{https://github.com/alexhe101/FourierISP}.
\end{abstract}
\section{Introduction}
The Image Signal Processing (ISP) pipeline, responsible for transforming RAW data captured by camera sensors into sRGB images, involves a series of low-level vision tasks including demosaicing, denoising, gamma correction, white balance, and color correction~\cite{Ramanath_Snyder_Yoo_Drew_2005}. Traditionally, these individual subprocesses are executed via independent algorithms, often requiring significant manual parameter adjustments~\cite{zhou2007image}. With the rise of mobile photography, smartphones have become the preferred choice for image capture due to their portability. 
However, the inherent limitations of sensor size and aperture in comparison to DSLR cameras pose challenges for mobile devices in achieving DSLR-like image quality. To bridge this gap, learning RAW-to-sRGB mapping through deep ISP models holds great promise~\cite{ignatov2020replacing,ignatov2020aim}.
These models hold the potential to convert mobile RAW data into high-quality sRGB images resembling those captured by DSLR cameras, without the need for manual fine-tuning.

Recent deep learning-based ISP methods that utilize mobile phone RAW images have shown promise in producing RGB images comparable to those from DSLR cameras~\cite{dai2020awnet,ignatov2021learned}. 
This innovation enables cost-effective mobile sensors to deliver visually appealing results. 
However, existing techniques often focus solely on the Raw-to-RGB conversion as a color mapping task, overlooking the crucial spatial relationship between different image types. This oversight leads to outputs with reduced clarity.
\begin{figure}
    \centering
\includegraphics[width=\linewidth]{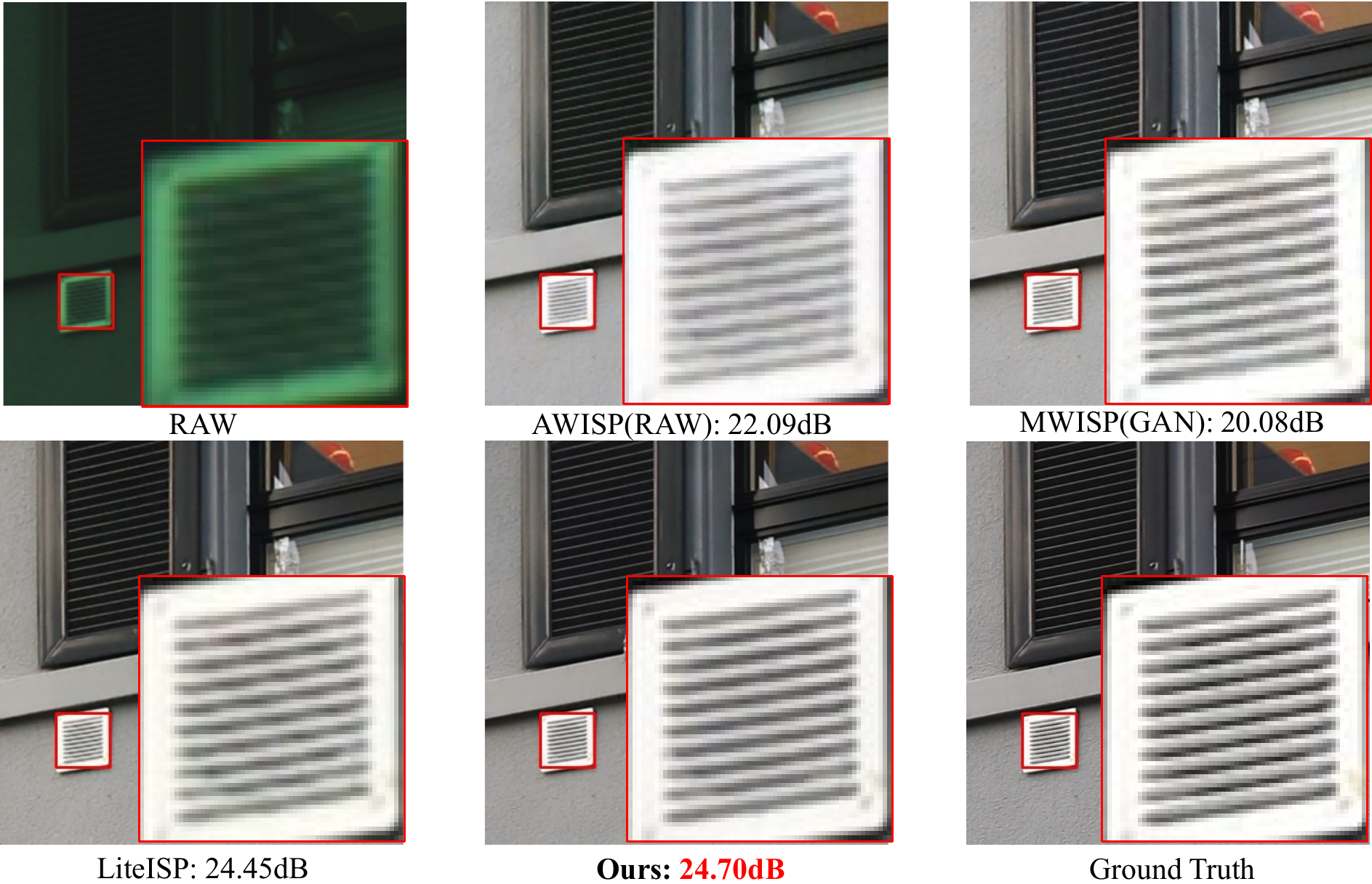}
    \caption{Results from the ZRR dataset. Our approach results in clear textures, surpassing other methods.}
    \label{fig:Motivation}
\end{figure}
In conventional ISP pipelines, processes like denoising, demosaicing, white balance, and gamma correction serve two interconnected yet distinct purposes: image restoration, which preserves spatial structure, and image enhancement, which involves color adjustments ~\cite{liang2021cameranet}. The effectiveness of enhancement relies heavily on successful restoration; any shortcomings in demosaicing or denoising can hinder the learning of color information.
Many widely-used models adopt one-stage design strategies that merge the learning of various ISP subtasks. Unfortunately, this approach limits the representation capability of CNN models and results in images lacking clear details and suffering from color distortions. Therefore, developing an efficient method to separate style and structural information is essential for unlocking enhanced performance within ISP frameworks.

\begin{figure}
    \centering
\includegraphics[width=\linewidth]{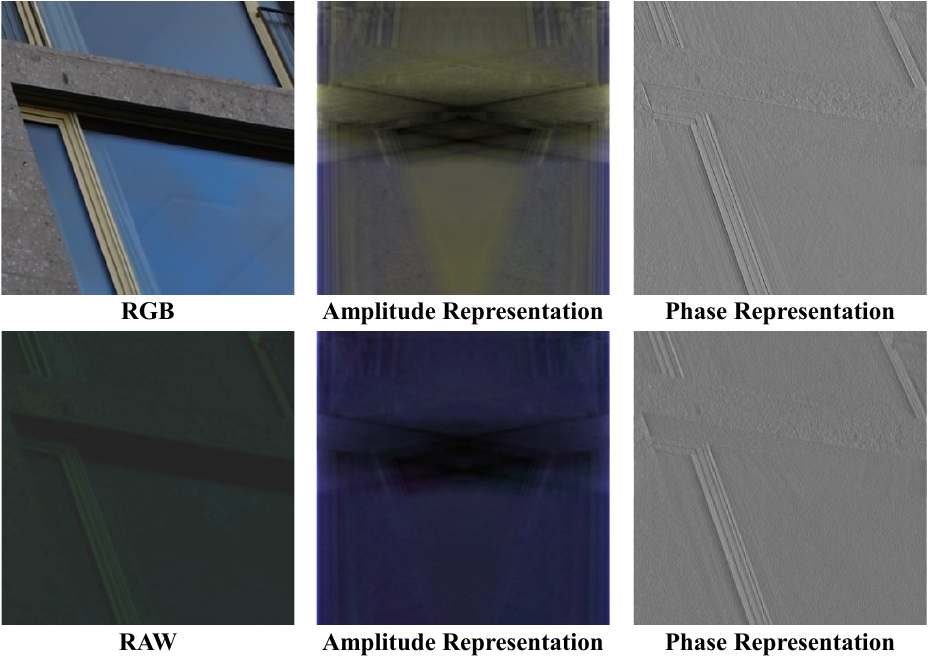}
    \caption{Illustration of image amplitude and phase representation. Clear distinction: phase captures spatial structure, while amplitude encodes color information.}
    \label{fig:Motivation}
\end{figure}

However, achieving effective style-structure decoupling in the spatial domain requires complex loss functions, which can be challenging~\cite{yang2023learning}. As a result, we shift our focus to the frequency domain, where the powerful prior of Fourier transform offers a promising solution for style and structure decoupling, as depicted in the Figure~\ref{fig:Motivation}. The Fourier transform of an image yields its amplitude, representing the style, and its phase, representing the structure. 
Notably, the phases of the two images are relatively close to each other, while the styles are quite different. 

Based on the aforementioned insights, we introduce a novel Neural ISP framework named Fourier-ISP.
Unlike previous methods, our approach utilizes the Fourier prior to decouple and optimize color knowledge and structure representation.
The network consists of three crucial subnets: the Phase Enhance Subnet (PES) for enhancing spatial structure and fine textures, the Amplitude Refine Subnet (ARS) for learning precise color information, and the Color Adaptation Subnet (CAS) responsible for transmitting color information to the phase-enhanced feature, thereby obtaining rich spatial details and ensuring precise color output. Through extensive evaluations on multiple datasets, our method showcases state-of-the-art results in qualitative and quantitative assessments, while also demonstrating robust transferability. 

Our contribution can be summarized as follows:
\begin{itemize}
    \item In this work, we present a novel approach utilizing the Fourier prior to decouple style and structure in the Raw-to-RGB mapping process. By separately optimizing the style and spatial structure of RAW images, we achieve highly accurate raw-to-RGB mapping results. 

    \item We propose the Fourier-ISP framework, consisting of three specialized sub-networks: PES, ARS, and CAS. This well-crafted architecture enables distinct subnets to acquire specific expertise, resulting in superior raw-to-RGB performance.

    \item The proposed approach outperforms the state-of-the-art on multiple datasets, as shown in extensive quantitative and qualitative experiments.  
\end{itemize}
\section{Related Work}
\subsection{Deep Learning for ISP}
In recent years, the integration of deep learning into ISP pipelines has gained significant attention due to the complexity manual adjustments in traditional pipelines. Some approaches target specific ISP modules, like image denoising~\cite{cheng2021nbnet}, demosaicing~\cite{liu2020joint}, and tone mapping~\cite{hu2022joint}, while others aim to overhaul the entire pipeline using neural network models. In prior works, both RGB and RAW images were acquired using the same device, as shown by DeepISP's~\cite{schwartz2018deepisp} end-to-end structure and CameraNet's~\cite{liang2021cameranet} division of the task into restoration and enhancement stages, guided by software-generated ground truth. Among recent endeavors, the Pynet and ZRRdataset~\cite{ignatov2020replacing} raise the challenging task of mapping mobile RAW images to DSLR camera RGB images, complicated by resolution discrepancies and spatial misalignment due to dual-device capture. Notable contributions include MW-ISPNet~\cite{ignatov2020aim} leveraging MWCNN~\cite{liu2018multi} for utilizing multi-scale features, AWNet~\cite{dai2020awnet} employing attention mechanisms for refined color learning and misalignment handling, and LiteISP's~\cite{zhang2021learning} lightweight design with optical flow alignment for fine texture output. 
LWISP~\cite{DBLP:conf/bmvc/ChenM22} adopt distillation for efficiency, achieving a parameter-effectiveness balance.
However, prevailing approaches treat RAW-to-RGB mapping as straightforward regression tasks, often ignoring disparities in structural and color information. While CameraNet recognizes these disparities, it requests intermediate result supervision from manual labeling. Our method stands apart by explicitly promoting style and color learning through frequency domain style-structure decoupling.

\subsection{Fourier Transform in Computer Vision}
In the realm of deep learning, the Fourier transform has attracted significant interest due to its distinctive attributes. Leveraging its global properties, FFC~\cite{chi2020fast} devised convolutional modules, effectively utilizing global information processing while maintaining minimal computational overhead. LAMA~\cite{suvorov2022resolution} employed FFC to construct an UHD image inpainting network by utilizing its global feature. 
In low-level vision, frequency domain attributes elevate high-frequency image details in super-resolution tasks~\cite{zhou2022spatial}, while image restoration tasks dissociate degradation features using Fourier transform~\cite{zhou2023fourmer}. However, its style-structure decoupling attributes have not been fully explored in the low level vision community and its application within the RAW-to-RGB remains uncharted.
\begin{figure*}[!h]
    \centering
    \includegraphics[width=\linewidth]{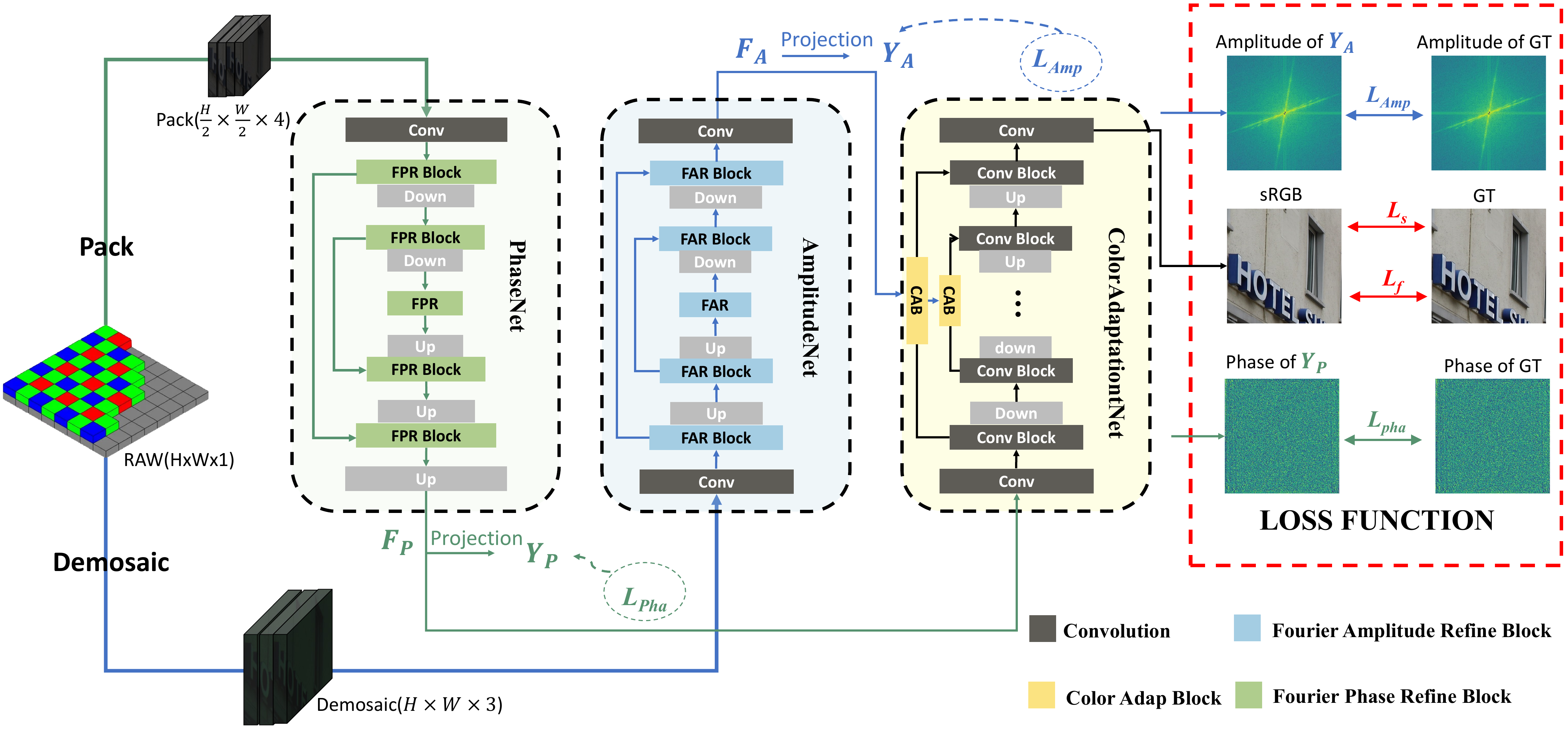}
    \caption{Our Model Framework. We employ separate processing for RAW images through packing and demosaicing. These processed images are subsequently fed into PES and ARS to learn the spatial details and style information of the image, respectively. Finally, we integrate the style information into the spatial features using the CAS and produce the final output.}
    \label{fig:mainfig}
\end{figure*}
\section{Method}
We leverage Fourier transform to decouple style and structure, enabling separate optimization of these two components. This section begins with an overview of the basic knowledge of Fourier transform, followed by an introduction of the network architecture and the loss functions.
\subsection{Fourier Transform of Image}
The Fourier Transform, widely employed in image processing, enables the conversion of signals into the frequency domain. The Fourier transform of an input image $x\in R^{h\times w}$  can be defined as follows:
\begin{equation}
\mathcal{F} (x)(u,v) = \frac{1}{\sqrt{HW} } \sum_{h=0}^{H-1} \sum_{w=0}^{W-1} x(h,w)e^{-j2\pi (\frac{h}{H}u+\frac{w}{W}v )}
\end{equation}
The amplitude and phase components are described as:
\begin{align}
        \mathcal{A} (x)(u,v) &= [R^2(x)(u,v)+I^2(x)(u,v)]^{\frac{1}{2} },\\
        \mathcal{P}(x)(u,v) &= \mathrm{arctan}[\frac{I(x)(u,v)}{R(x)(u,v)}]
\end{align}
where $I(x)$ and $R(x)$ indicate imaginary and real parts of the image's frequency representation $\mathcal{F} (x)$, correspondingly.

The Fourier transform offers two distinctive attributes. Firstly, its global nature arises from the weighted summation of spatial domain values, thus inherently containing global information. By leveraging this, we go beyond the limitations of local receptive fields in convolutional neural networks, acquiring pivotal global context for effective color mapping. Furthermore, the Fourier transform yields amplitude and phase in the frequency domain. The former signifies style, while the latter encapsulates structure. Noise in raw images commonly resides in the phase component~\cite{DBLP:conf/iclr/LiGZLZFL23}, thus preliminary denoising aids the network in grasping color details more effectively. Fourier transform's decoupling facet enabling us separate processing of these distinct information.
\subsection{Network Framework}
The network architecture, depicted in Figure~\ref{fig:mainfig}, processes the RAW image $\mathbf{R}^{ H\times W\times 1}$. $\mathbf{R}$ undergoes packing and demosaicing to yield $\mathbf{R_{pack}}^{\frac{H}{2}\times \frac{W}{2} \times 4}$ and $\mathbf{R_{dem}}^{H\times W\times 3}$, respectively. PES, ARS, and CAS form the core components.

PES processes $\mathbf{R_{pack}}$ for spatial structure enhancement, yielding refined feature $\mathbf{F_P}^{H \times W \times C}$. The projection of $\mathbf{F_p}$ into RGB space involves convolutional steps, supervised with phase from the ground truth (GT) image. Simultaneously, $\mathbf{R_{dem}}$ is directed to ARS, optimizing amplitude to generate $\mathbf{F_A}^{H \times W \times C}$. This amplitude-focused information is then projected back into RGB space through convolutional projection, with supervision from GT image amplitudes.

Next, $\mathbf{F_P}$ is fed into CAS, leveraging $\mathbf{F_A}$'s amplitude to adjust color information via color adaptation blocks, as amplitude encodes image style. CAS generates the output image. Spatial loss and frequency domain loss govern final result supervision.
\begin{figure}
    \centering
    \includegraphics[width=\linewidth]{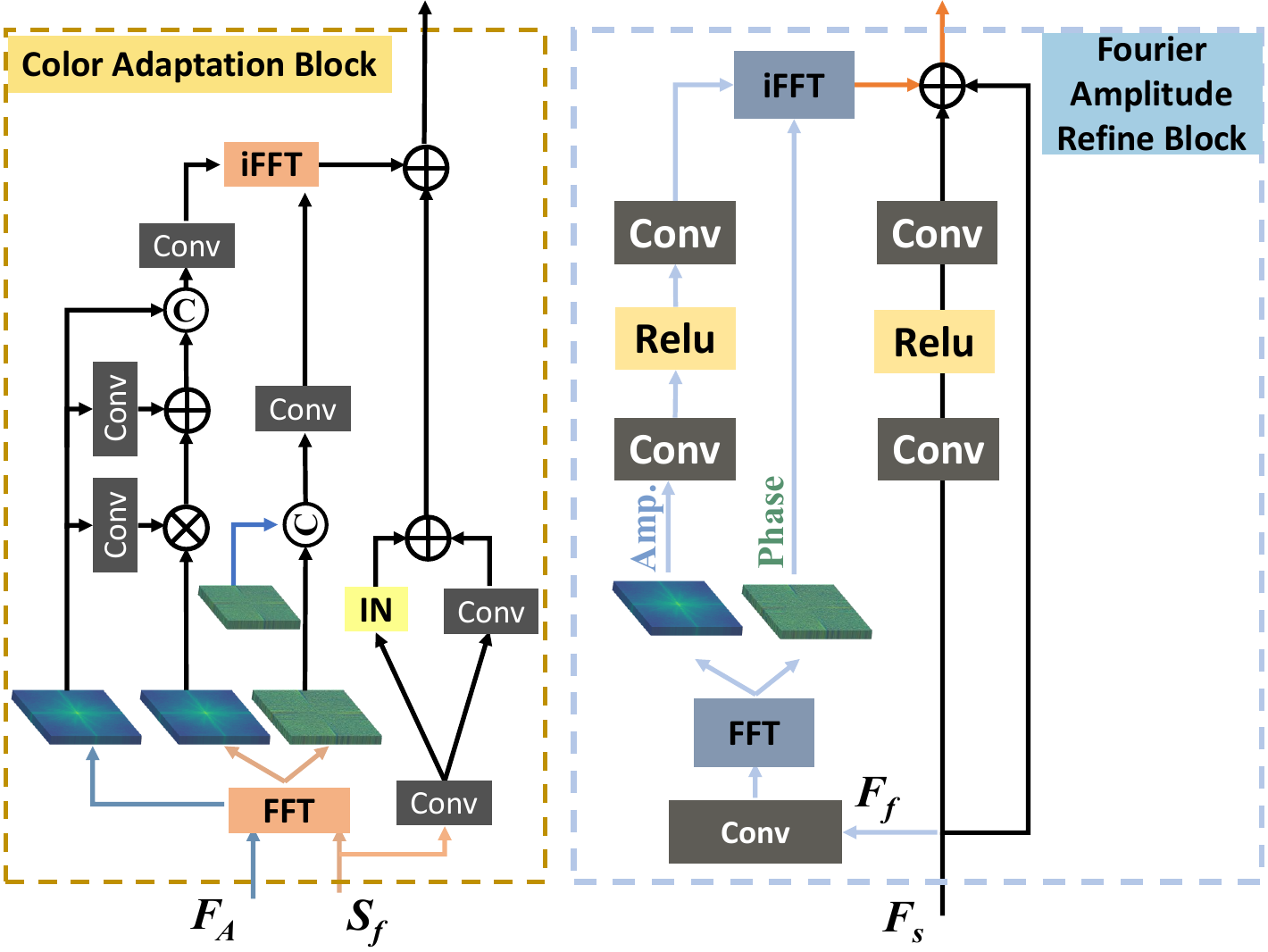}
    \caption{The left portion of the figure illustrates the Color Adaptation Block, while the right side showcases the Fourier Amplitude Refine Block.}
    \label{fig:block}
\end{figure}
\subsection{Key Components.}
\noindent\textbf{Phase Enhance Subnet.}
The PES is specifically designed to enhance the spatial structure of the input images. Its primary function is to align the image phase with the ground truth, which facilitates fine-grained spatial structure learning. By optimizing the spatial structure, we can effectively reduce noise in the final output and generate finer image textures since image noise predominantly resides within the phase components. We start by packing the RAW image into four channels, a preprocessing step that enhances the image's suitability for spatial structure learning~\cite{dai2020awnet}. The input to PES, denoted as $\mathbf{R_{pack}}$, undergoes processing, and we perform upsampling through PixelShuffle at the end.
The process of PES can be described as follows:
\begin{align}
    &\mathbf{F_p} = \phi(\mathbf{R_{pack}}),\\
    &\mathbf{Y_p} = Proj(\mathbf{F_p})
\end{align}
Here, $Proj(.)$ represents the $1\times1$ convolution operator that projects $\mathbf{F_p}$ back to the RGB domain, and $\phi(.)$ corresponds to the PES.

The core component of PES is the Fourier Phase Refine Block (FPRB), depicted in Figure~\ref{fig:block}. It's worth noting that the main distinction between FPRB and the Fourier Amplitude Refine Block (FARB) is that FPRB deals with the phase, while FARB focuses on the amplitude. This module utilize the interaction between image phase information and spatial features, leading to efficient extraction of complex image details. 

The operations within the FARB can be defined as follows, given the input features $\mathbf{F_f}$ and $\mathbf{F_s}$:
\begin{align}
    &\mathbf{F^{1}_f} = Conv(\mathbf{F_f}),\\
    &\mathcal{A}(\mathbf{F^{1}_f}),\mathcal{P}(\mathbf{F^{1}_f}) = \mathcal{F}(\mathbf{F^{1}_f}),\\
    &\mathbf{F^{2}_f} = \mathcal{F}^{-1}(Conv(\mathcal{A}(\mathbf{F^{1}_f})),\mathcal{P}(\mathbf{F^{1}_f})),\\ 
    &\mathbf{F_{out}} = \mathbf{F^{2}_f}+Conv(\mathbf{F_s})+\mathbf{F_s}.
\end{align}
\noindent\textbf{Amplitude Refine Subnet.}
The architecture of the ARS closely resembles that of PES. ARS uses $\mathbf{R_{dem}}$ as input, ensuring that images of the same size are better suited for learning color mapping. In ARS, the primary objective is to optimize the image's amplitude to align it with the GT image. This amplitude component encodes the style of the image and encompasses global information, making color learning more robust to dataset misalignment. The utilization of global information is crucial for effective color learning, as images color are influenced by both local details and global features. Optimizing amplitude for color learning allows us to leverage the global attributes of the Fourier transform, enabling the capture of essential global information.
.
The FARB, as depicted in Figure~\ref{fig:block}, plays a central role in ARS for optimizing image amplitude. Its core structure is similar to the FPRB, with the key distinction being that FARB focuses on optimizing amplitude instead of phase.

The ARS process can be concisely expressed as follows:
\begin{align}
    &\mathbf{F_A} = \gamma(\mathbf{R_{dem}}),\\
    &\mathbf{\mathbf{Y_A}} = Proj(\mathbf{F_A})
\end{align}
Here, $Proj(.)$ symbolizes a $1\times1$ convolution operator, projecting $F_A$ back into the RGB domain and $\gamma(.)$ is ARS.

\noindent\textbf{Color Adaptation SubNet.}
CAS serves as a pivotal component with the primary aim of fusing the refined spatial structures with accurate color information, thereby enabling precise adjustments to the color attributes within the feature maps. This sub-network adopts a multi-scale approach, injecting color features into the feature maps at different scales, resulting in targeted enhancement of both the overall color information and spatial structure across the entire image.
Within CAS, a straightforward Unet architecture is employed, complemented by a custom-designed Color Adaptation Block (CAB), depicted in Figure~\ref{fig:block}, to modulate the feature map. We utilize amplitude features to modulate the style information within the feature map for color information is predominantly encoded by amplitude.

In CAB, the feature map is split into two branches. In the frequency branch, we apply Fourier transform to $\mathbf{F_A}$ and $\mathbf{S_f}$ independently. The amplitude of $\mathbf{F_A}$ is used to modulate the amplitude of $\mathbf{S_f}$ using the SFT~\cite{wang2018recovering} mechanism, and the phase components from both features are simply fused. After performing inverse Fourier transform, we obtain the frequency features. In the spatial domain branch, we employ the HINBlock~\cite{chen2021hinet} for efficient feature extraction, and subsequently, sum up the spatial and frequency features to complete the color adaptation process. The adapted feature map is then downsampled, serving as the $\mathbf{F_A}$ for the next scale.
\subsection{Loss Function}
Our comprehensive loss function comprises three essential components: Phase loss for PES, Amplitude loss for ARS, and the combined spatial and frequency loss for the final output.

PES and ARS are supervised using Phase Loss and Amplitude Loss. The Phase Loss $\mathcal{L}_{pha}$ is computed as the L1 norm between the phase $\mathcal{P}(\mathbf{Y_P})$ of the $\mathbf{Y_P}$ and the reference image GT's phase $\mathcal{P}(\mathbf{G})$. Similarly, the Amplitude Loss $\mathcal{L}_{amp}$ is calculated as the L1 norm between the amplitude $\mathcal{A}(\mathbf{Y_A})$ of $\mathbf{\mathbf{Y_A}}$ and the reference image GT's amplitude $\mathcal{A}(\mathbf{G})$.
\begin{align}
    &\mathcal{L}_{pha} = ||\mathcal{P}(\mathbf{Y_P})-\mathcal{P}(\mathbf{G})||_{1},\\
    &\mathcal{L}_{amp} = ||\mathcal{A}(\mathbf{Y_A})-\mathcal{A}(\mathbf{G})||_{1}.
\end{align}
The final network output, denoted as $\mathbf{Y}$, is constrained by $L_{spa}$ (spatial loss) and $L_{fre}$ (frequency domain loss).
Regarding the spatial loss, we utilize a combined approach involving VGG~\cite{johnson2016perceptual}, SSIM, and L1 losses. In the context of the frequency domain loss, we optimize the real and imaginary components of both $\mathbf{Y}$ and the reference image $\mathbf{G}$ after performing Fourier transform, which we represent as $R(Y)$, $R(G)$, $I(Y)$, and $I(G)$.
The formulation of this loss component can be briefly expressed as follows:
\begin{align}
    &\mathcal{L}_{spa} = \mathcal{L}_{vgg}+0.5*\mathcal{L}_{ssim}+\mathcal{L}_1,\\
    &\mathcal{L}_{fre} = ||R(Y)-R(G)||_1+||I(Y)-I(G)||_1
\end{align}
Our comprehensive loss function is a weighted summation of the components outlined above:
\begin{align}
\mathcal{L}_{total} = \mathcal{L}_{spa} + \alpha * \mathcal{L}_{fre} + \beta * \mathcal{L}_{pha} + \gamma * \mathcal{L}_{amp}
\end{align}
In our implementation, we set the weights $\alpha$, $\beta$, and $\gamma$ to 0.1 based on experience. 
Due to our incorporation of global information within the losses, our approach demonstrates reduced sensitivity to dataset misalignment issues.

\section{Experiment}
\begin{table*}[!ht]
\begin{center}
    
\scalebox{0.75}{
\begin{tabular}{c|cccc|cccc|cccc}
\hline
\multirow{2}{*}{Methods} & \multicolumn{4}{c|}{ZRR}                                                                        & \multicolumn{4}{c|}{ZRR(Align GT with RAW)}                                                     & \multicolumn{4}{c}{MAI}                                                                         \\ \cline{2-13} 
                         & \multicolumn{1}{c|}{PSNR$\uparrow$}  & \multicolumn{1}{c|}{SSIM$\uparrow$}   & \multicolumn{1}{c|}{MS-SSIM$\uparrow$} & LPIPS$\downarrow$ & \multicolumn{1}{c|}{PSNR$\uparrow$}  & \multicolumn{1}{c|}{SSIM$\uparrow$}   & \multicolumn{1}{c|}{MS-SSIM$\uparrow$} & LPIPS$\downarrow$ & \multicolumn{1}{c|}{PSNR$\uparrow$}  & \multicolumn{1}{c|}{SSIM$\uparrow$}   & \multicolumn{1}{c|}{MS-SSIM$\uparrow$} & LPIPS$\downarrow$ \\ \hline
PyNet                    & \multicolumn{1}{c|}{21.19} & \multicolumn{1}{c|}{0.7471} & \multicolumn{1}{c|}{0.8620}  & 0.1930 & \multicolumn{1}{c|}{22.73} & \multicolumn{1}{c|}{0.8451} & \multicolumn{1}{c|}{/}       & 0.1520 & \multicolumn{1}{c|}{/}     & \multicolumn{1}{c|}{/}      & \multicolumn{1}{c|}{/}       & /     \\
AWNET(raw)               & \multicolumn{1}{c|}{21.42} & \multicolumn{1}{c|}{0.7478} & \multicolumn{1}{c|}{0.8609}  & 0.1980 & \multicolumn{1}{c|}{23.27} & \multicolumn{1}{c|}{0.8542} & \multicolumn{1}{c|}{0.9312}  & 0.1510 & \multicolumn{1}{c|}{23.95} & \multicolumn{1}{c|}{0.8583} & \multicolumn{1}{c|}{0.9508}  & 0.1640 \\
AWNET(demosaic)          & \multicolumn{1}{c|}{21.53} & \multicolumn{1}{c|}{0.7488} & \multicolumn{1}{c|}{0.8614}  & 0.2120 & \multicolumn{1}{c|}{23.38} & \multicolumn{1}{c|}{0.8497} & \multicolumn{1}{c|}{0.9297}  & 0.1640 & \multicolumn{1}{c|}{24.03} & \multicolumn{1}{c|}{0.8670} & \multicolumn{1}{c|}{0.9525}  & 0.1340 \\
MWISP                    & \multicolumn{1}{c|}{21.42} & \multicolumn{1}{c|}{0.7544} & \multicolumn{1}{c|}{0.8654}  & 0.2130 & \multicolumn{1}{c|}{23.07} & \multicolumn{1}{c|}{0.8479} & \multicolumn{1}{c|}{0.9255}  & 0.1650 & \multicolumn{1}{c|}{24.24} & \multicolumn{1}{c|}{0.8558} & \multicolumn{1}{c|}{0.9491}  & 0.1280 \\
MWISP-GAN                & \multicolumn{1}{c|}{21.16} & \multicolumn{1}{c|}{0.7317} & \multicolumn{1}{c|}{0.8578}  & \textbf{0.1580} & \multicolumn{1}{c|}{22.80} & \multicolumn{1}{c|}{0.8285} & \multicolumn{1}{c|}{0.9234}  & 0.1340 & \multicolumn{1}{c|}{24.34} & \multicolumn{1}{c|}{0.8568} & \multicolumn{1}{c|}{0.9234}  & 0.1150 \\
LiteISP                  & \multicolumn{1}{c|}{21.55} & \multicolumn{1}{c|}{0.7487} & \multicolumn{1}{c|}{0.8596}  & 0.1870 & \multicolumn{1}{c|}{23.76} & \multicolumn{1}{c|}{0.8730} & \multicolumn{1}{c|}{0.9450}  & 0.1330 & \multicolumn{1}{c|}{23.11} & \multicolumn{1}{c|}{0.7941} & \multicolumn{1}{c|}{0.9250}  & 0.1990 \\
LWISP                    & \multicolumn{1}{c|}{21.57} & \multicolumn{1}{c|}{/}      & \multicolumn{1}{c|}{0.8622}  & /     & \multicolumn{1}{c|}{/}     & \multicolumn{1}{c|}{/}      & \multicolumn{1}{c|}{/}       & /     & \multicolumn{1}{c|}{/}     & \multicolumn{1}{c|}{/}      & \multicolumn{1}{c|}{/}       & /     \\ \hline
Ours                     & \multicolumn{1}{c|}{\textbf{21.65}} & \multicolumn{1}{c|}{\textbf{0.7546}} & \multicolumn{1}{c|}{\textbf{0.8660}}  & 0.1820 & \multicolumn{1}{c|}{\textbf{23.93}} & \multicolumn{1}{c|}{\textbf{0.8744}} & \multicolumn{1}{c|}{\textbf{0.9461}}  & \textbf{0.1240} & \multicolumn{1}{c|}{\textbf{24.99}} & \multicolumn{1}{c|}{\textbf{0.8820}} & \multicolumn{1}{c|}{\textbf{0.9594}}  & \textbf{0.0850} \\ \hline
\end{tabular}

}
 		\caption{Quantitative comparison on three datasets. Best results are highlighted by \textbf{bold}. $\uparrow$ indicates that the larger the value, the better the performance, and $\downarrow$ indicates that the smaller the value, the better the performance.}
\label{camp}
\end{center} 

\end{table*}
\subsection{Datasets and Benchmark}
We conducted evaluations on two distinct datasets: the ZRR dataset and the MAI dataset~\cite{ignatov2021learned}. The ZRR dataset involves mapping RAW images from the Huawei P20 camera to RGB images from a Canon camera. Meanwhile, the MAI dataset focuses on mapping Sony IMX586 Quad Bayer RAW images to Fuji camera RGB images. Notably, the RAW images in the ZRR dataset possess a bit width of 10 bits, whereas the MAI dataset RAW images have a bit width of 12 bits. The selection of the latter dataset allows us to assess the model's transferability.
Given the substantial dissimilarity between the RAW images in these two datasets, we initially train the model on the ZRRdataset. To evaluate the model's transferability, we perform a one-epoch fine-tuning process on the MAI dataset. This approach serves to gauge how well the model can adapt to the differences in RAW image characteristics.

In our comparative analysis, we include state-of-the-art methods such as Pynet, AWNet, MWISP, MWISPGAN, LiteISP, and LWISP. We utilize reference evaluation metrics, including PSNR, SSIM, MS-SSIM~\cite{wang2003multiscale}, and LPIPS~\cite{zhang2018unreasonable}. 
For more experiments results, please refer to the supplementary material
\subsection{Implementation details}
We conducted our experiments utilizing the PyTorch framework on four Titan XP GPUs, encompassing a total of $3*10^{4}$ training iterations. Employing the Adam optimizer, we initially set the learning rate at $2*10^{-4}$, progressively halving it at every $1*10^{4}$ iterations to fine-tune the training process. Notably, for the ZRRdataset, we utilized a patch size of 448x488 for both training and testing. In contrast, for the MAI dataset, the patch size was set to 224x224 for the same purposes.
\begin{figure}[]
    \centering
    \includegraphics[width=\linewidth]{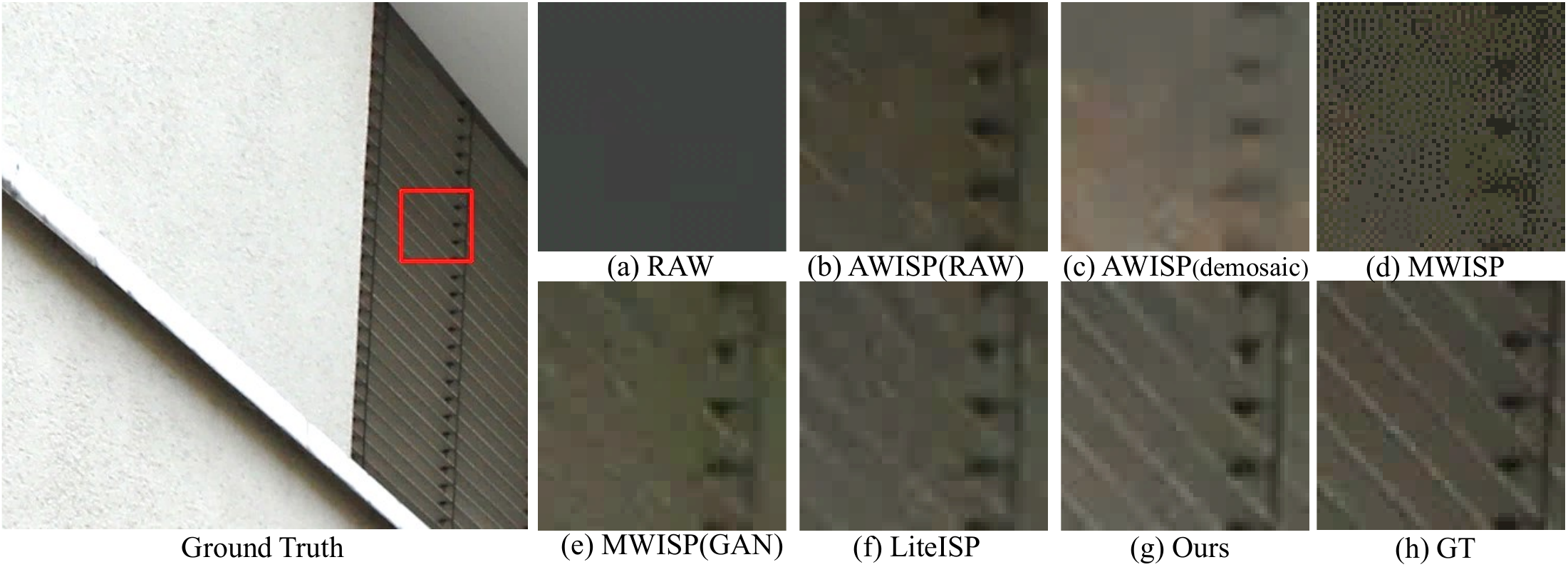}
    \caption{Experimental results from the ZRRdataset. our approach excels in capturing intricate texture details. RAW images are difficult to visualize due to dark local areas.
}
    \label{fig:details}
\end{figure}
\subsection{Comparison with state-of-the-art methods}  
\noindent\textbf{Evaluation on Quantitative metric.} 
We conducted a comprehensive comparison of our proposed method against the SOTA approaches on both the ZRR and MAI datasets, as illustrated in the Table~\ref{camp}. For the MAI dataset, we employed a pre-trained model on the ZRR dataset and performed one epoch of fine-tuning for direct comparison. Additionaly,acknowledging the partial misalignment between the ground truth and input in the ZRR dataset, we employed the optical flow network~\cite{sun2018pwc} from LiteISP to align the test set and calculate the evaluation metrics. Due to unavailability of pretrained weights of Pynet and source code of LWISP, we based our evaluation of both methods solely on the metrics provided in their respective publication.

Our method achieved PSNR improvement of 0.08dB compared to the SOTA method on the ZRR dataset. Additionally, our method exhibited advancements in SSIM and LPIPS metrics. Remarkably, on the aligned dataset, our method achieved a noteworthy PSNR improvement of 0.17dB. This further reinforces the effectiveness of our style and structure decoupling approach and demonstrates the robustness of the global loss in handling data misalignment.

Moreover, our model exhibits strong transferability, surpassing the performance metrics of other methods on the MAI dataset. It is notable that the requirement for an additional optical flow estimation network during LiteISP training limits the transferability of the model, leading to bad result on the MAI dataset.

\begin{figure*}[!h]
    \centering
    \includegraphics[width=\linewidth]{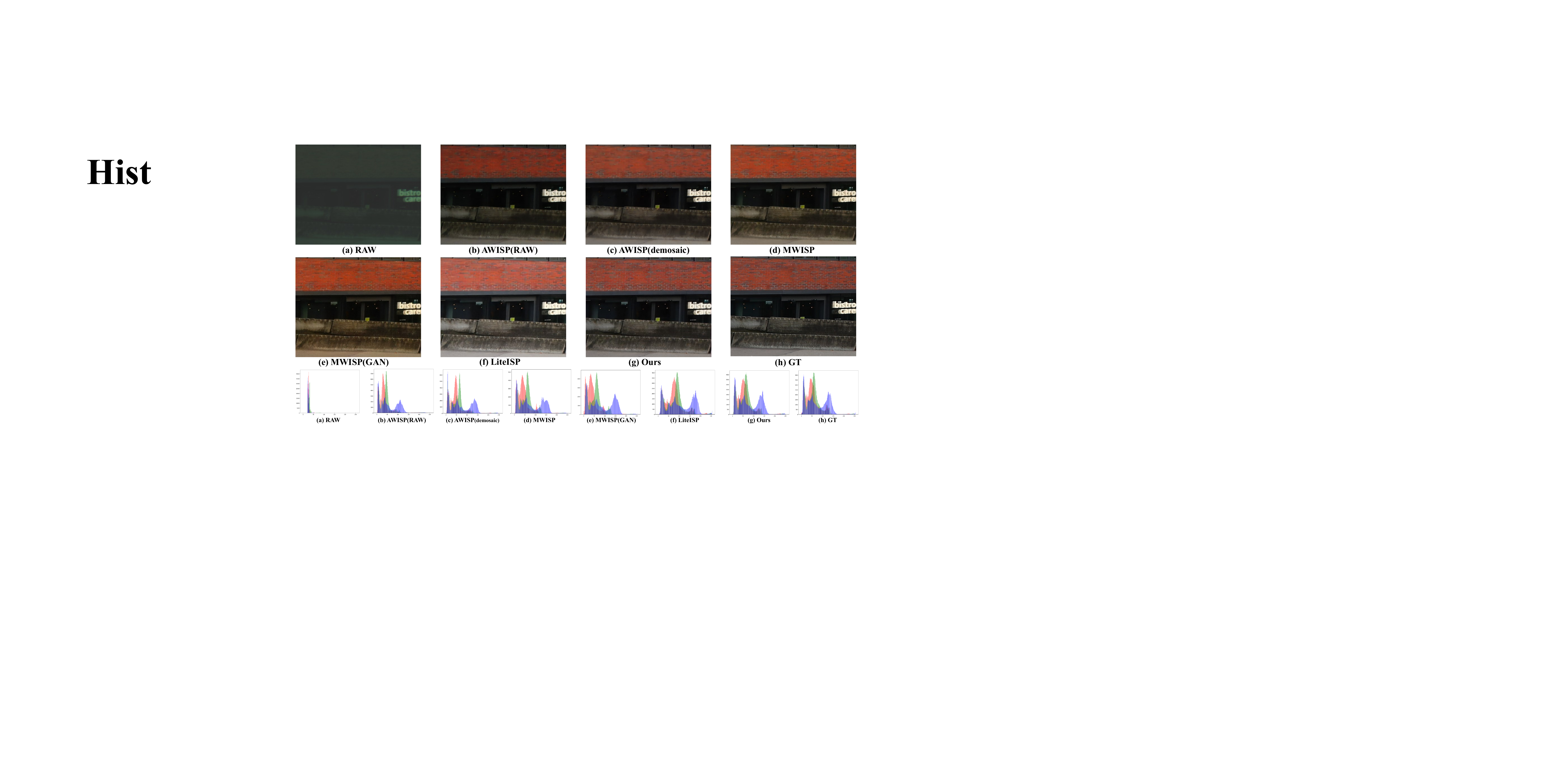}
    \caption{The results image from ZRRdataset. The last row showcase the color histogram of the image.
Our method has the closest color to GT image.}
    \label{fig:color}
\end{figure*}
\begin{figure*}[!h]
    \centering
    \includegraphics[width=\linewidth]{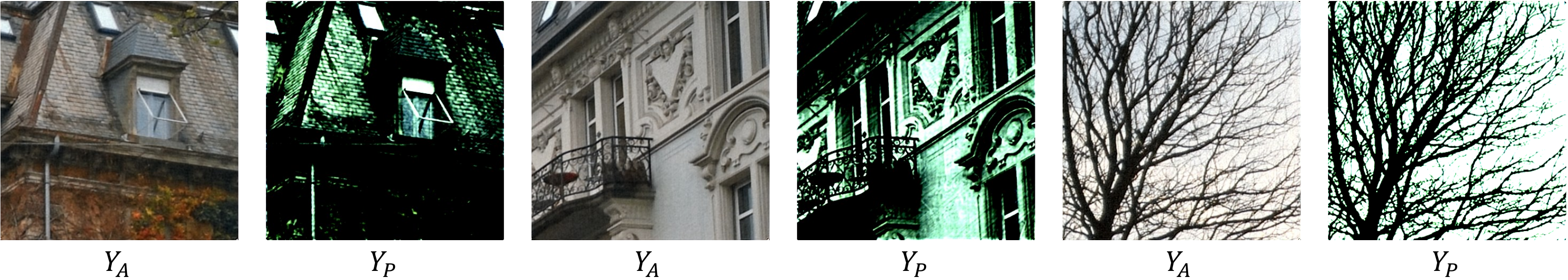}
    \caption{The Amplitude and Phase feature map of our methods.}
    \label{fig:fmap}
\end{figure*}
\noindent\textbf{Evaluation on Qualitative metric.} 
\begin{figure}[!h]
    \centering
    \includegraphics[width=\linewidth]{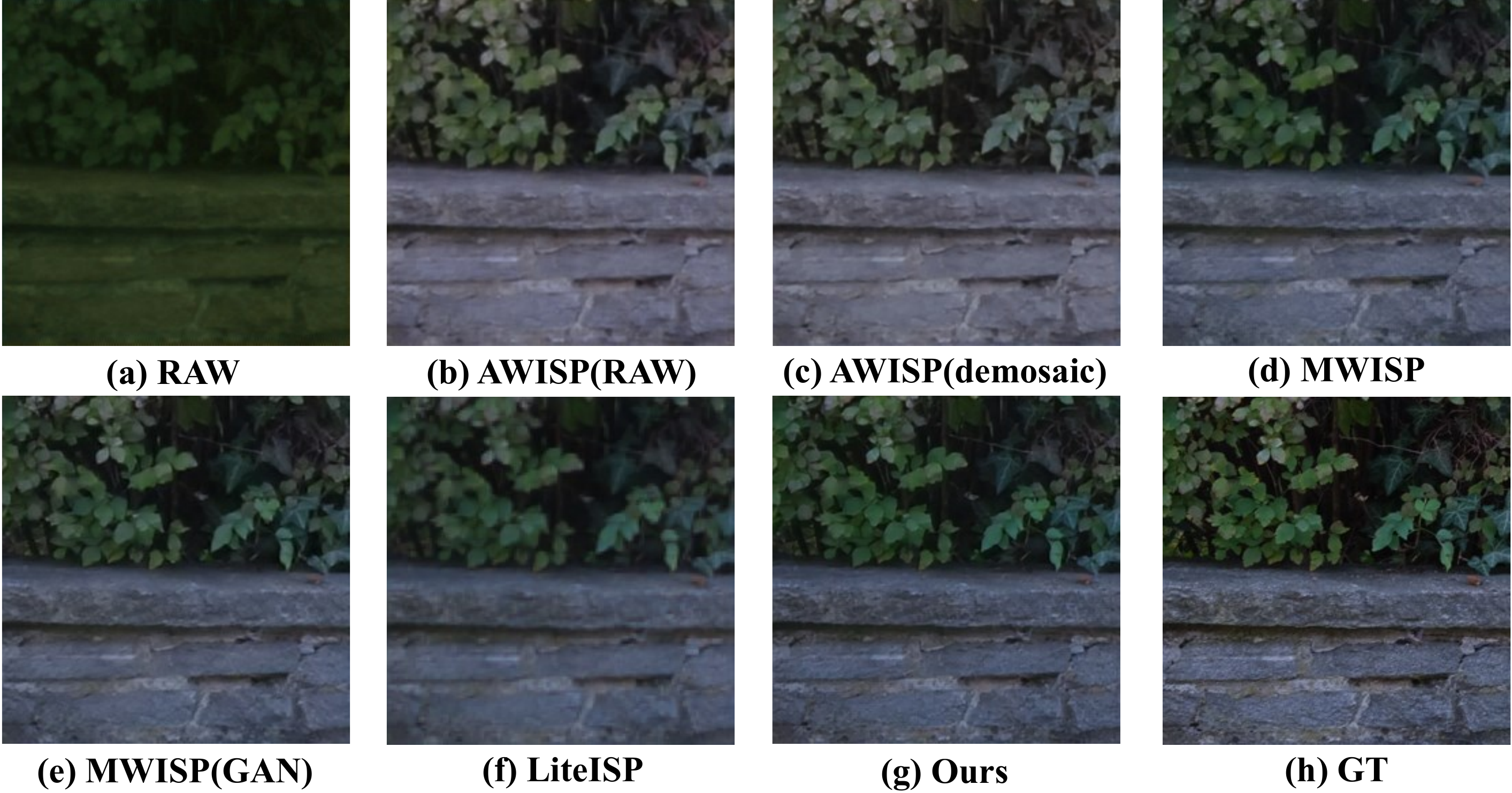}
    \caption{MAI dataset result image. Our method showcases the superiortransferability, closely aligned with the Ground Truth.}
    \label{fig:MAI}
\end{figure}

In our qualitative experiments, we select three typical images from two datasets to comprehensively showcase the effectiveness of our method across three crucial aspects: image texture, color fidelity, and model transfer capability. As illustrated in Figure~\ref{fig:details}, we magnified local textures from the ground truth to emphasize the detail-capturing capabilities of various models. The visual comparison clearly indicates that our method, leveraging fine processing of spatial structures and the capacity to capture high-frequency information in the frequency domain, consistently generates more refined results, in contrast to other methods that struggle to capture such textures, even leading to artifacts.

In Figure~\ref{fig:color}, we present a comprehensive comparison of color fidelity, along with accompanying color histograms for each image. Analyzing the histograms reveals that our method's results closely align with the ground truth, signifying our superior color reproduction. Finally, we demonstrate the model's transferability by comparing images from the MAI dataset in Figure~\ref{fig:MAI}. The comparison highlights that alternative methods exhibit incorrect color mappings or spatial structures, while our method consistently aligns more closely with the ground truth in terms of both color and spatial information. 

\subsection{Ablation Experiments}
We conducted multiple ablation experiments on the ZRR dataset to validate our method, and conducted experiments on multiple dimensions such as model structure, loss function, and model parameter quantity.

\noindent\textbf{Model structure.}
Our model is thoughtfully designed to capture both spatial structures and style features, with dedicated modules in PES and ARS, respectively. In this ablation experiments, we removed certain components from the network's architecture. Specifically, in the PES configuration, we omitted the frequency branch of FPRB and the corresponding phase loss, and similarly, in ARS, we excluded the frequency branch of FARB, the CAB, and the Amplitude Loss. We directly concatenate $\mathbf{F_A}$ with $\mathbf{F_P}$ into CAS. The results of ablation experiments are presented in the first and second rows of the Table~\ref{abl}. Removal of any module leads to a significant drop in both evaluation metrics. Specifically, when Phase information is omitted, SSIM indicators decrease notably, indicating severe damage to the result image spatial structure. Conversely, the removal of Amplitude information primarily affects the PSNR indicator.

\noindent\textbf{Loss Function.} In the examination of the loss function, we conducted empirical investigations by substituting the AmplitudeLoss in AmplitudeNet with ColorHistLoss to assess its influence on the final results. Prior research~\cite{afifi2021histogan} has utilized ColorHistLoss to approximate the color distribution of an image towards a reference image. Our findings shown in the third row of Talbe~\ref{abl} indicate that when replacing AmplitudeLoss with ColorHistLoss, the model's performance indicators exhibited a decline, emphasizing that the coarse-grained histogram partitioning approach is not well-suited for the Raw to RGB task.

\noindent\textbf{Model parameters.} In addition, we conducted a comprehensive investigation into the influence of parameter quantities on model performance. Our initial experiment employed a base channel count of 24, while in the ablation study, we explored channel counts of 16 and 48 to assess their impact on model effectiveness. As depicted in the results table~\ref{abl2}, a direct correlation is observed between the number of parameters and the model's performance, with larger parameter counts leading to improved outcomes. While constraining the channel count to 16 effectively reduces the model's parameter count, and the achieved PSNR index remains comparable to our existing methods, the visual results fall short of the quality attained with our current settings. 
our model possesses half the number of parameters compared to LiteISP and only one-sixth of MWISP, placing it on par with LWISP. This highlights our model's equilibrium between the quantity of model parameters and evaluation metrics.

\begin{table}[]
	\centering
\scalebox{0.8}{
\begin{tabular}{c|cccc}
\hline
\multirow{2}{*}{Configuration} & \multicolumn{4}{c}{ZRR}                                                                         \\ \cline{2-5} 
                               & \multicolumn{1}{c|}{PSNR}  & \multicolumn{1}{c|}{SSIM}   & \multicolumn{1}{c|}{MS-SSIM} & LPIPS \\ \hline
FourierISP w/o Phase           & \multicolumn{1}{c|}{21.47} & \multicolumn{1}{c|}{0.7413} & \multicolumn{1}{c|}{0.8581}  & 0.199 \\
FourierISP w/o Amplitude       & \multicolumn{1}{c|}{21.37} & \multicolumn{1}{c|}{0.7488} & \multicolumn{1}{c|}{0.8592}  & 0.187 \\
FourierISP w/ ColorHist        & \multicolumn{1}{c|}{21.13} & \multicolumn{1}{c|}{0.7476} & \multicolumn{1}{c|}{0.8581}  & 0.193 \\ \hline
FourierISP                     & \multicolumn{1}{c|}{\textbf{21.65}} & \multicolumn{1}{c|}{\textbf{0.7546}} & \multicolumn{1}{c|}{\textbf{0.8660}}  & \textbf{0.182} \\ \hline
\end{tabular}}
 		\caption{The results of the ablation experiments conducted on the ZRRdataset}
   \label{abl}

\end{table}
\begin{table}[]
	\centering
\scalebox{0.7}{
\begin{tabular}{c|c|cccc}
\hline
\multirow{2}{*}{Configuration} & \multirow{2}{*}{Params(M)} & \multicolumn{4}{c}{ZRR}                                                                         \\ \cline{3-6} 
                               &                            & \multicolumn{1}{c|}{PSNR}  & \multicolumn{1}{c|}{SSIM}   & \multicolumn{1}{c|}{MS-SSIM} & LPIPS \\ \hline
FourierISP 16 channels         & 2.75                       & \multicolumn{1}{c|}{21.64} & \multicolumn{1}{c|}{0.7534} & \multicolumn{1}{c|}{0.8609}  &{0.186} \\
FourierISP 48 channels         & 24.6                       & \multicolumn{1}{c|}{\textbf{21.67}} & \multicolumn{1}{c|}{\textbf{0.7548}} & \multicolumn{1}{c|}{\textbf{0.8669}}  &  \textbf{0.180} \\
FourierISP                     & 6.17                       & \multicolumn{1}{c|}{21.65} & \multicolumn{1}{c|}{0.7546} & \multicolumn{1}{c|}{0.8660}  & {0.182} \\ \hline
\end{tabular}}
 		\caption{The results of the ablation experiments conducted on the ZRRdataset}
   \label{abl2}

\end{table}
\subsection{Visualization of Feature Maps}
To demonstrate the distinctive capabilities of our subnetworks, we conducted feature map visualizations shown in Figure~\ref{fig:fmap}. Specifically, we projected $\mathbf{F_A}$ and $\mathbf{F_P}$ into the RGB space, yielding $\mathbf{Y_A}$ and $\mathbf{Y_P}$, respectively. The visual analysis reveals that PES adeptly captured the spatial structure, while ARS faithfully preserved the image's style attributes.

Remarkably, $\mathbf{Y_A}$ captures the overall style of the image. However, it tends to lack granularity in representing local details. On the other hand, $\mathbf{Y_P}$ focuses on preserving the structural details and finer local information of the image but lacks color information.

This visualization underscores the effectiveness of our method in handling both structural and style information.
\section{Conclusion}
This paper introduces a novel approach for RAW to RGB mapping, leveraging the power of Fourier transform to disentangle image style and structure. By processing these two aspects independently, our method achieves remarkable accuracy in color reproduction and texture preservation. Our proposed Fourier ISP framework features Phase Enhance Networks, Amplitude Refine Networks, and Color Adaptation Networks, enabling separate learning of style and structure, followed by a coherent integration for the final output. Extensive quantitative and qualitative experiments on multiple datasets demonstrate the superiority of our approach over state-of-the-art methods.
\section{Acknowledgement}
This work was Supported by the Natural Science Foundation of Anhui Province (No.2208085MC57), and HFIPS Director’s Fund, Grant No.2023YZGH04.

\bibliography{aaai24}

\begin{thebibliography}{27}
\providecommand{\natexlab}[1]{#1}

\bibitem[{Afifi, Brubaker, and Brown(2021)}]{afifi2021histogan}
Afifi, M.; Brubaker, M.~A.; and Brown, M.~S. 2021.
\newblock Histogan: Controlling colors of gan-generated and real images via color histograms.
\newblock In \emph{Proceedings of the IEEE/CVF conference on computer vision and pattern recognition}, 7941--7950.

\bibitem[{Chen and Ma(2022)}]{DBLP:conf/bmvc/ChenM22}
Chen, H.; and Ma, K. 2022.
\newblock {LW-ISP:} {A} Lightweight Model with {ISP} and Deep Learning.
\newblock In \emph{33rd British Machine Vision Conference 2022, {BMVC} 2022, London, UK, November 21-24, 2022}, 148. {BMVA} Press.

\bibitem[{Chen et~al.(2021)Chen, Lu, Zhang, Chu, and Chen}]{chen2021hinet}
Chen, L.; Lu, X.; Zhang, J.; Chu, X.; and Chen, C. 2021.
\newblock Hinet: Half instance normalization network for image restoration.
\newblock In \emph{Proceedings of the IEEE/CVF Conference on Computer Vision and Pattern Recognition}, 182--192.

\bibitem[{Cheng et~al.(2021)Cheng, Wang, Huang, Liu, Fan, and Liu}]{cheng2021nbnet}
Cheng, S.; Wang, Y.; Huang, H.; Liu, D.; Fan, H.; and Liu, S. 2021.
\newblock Nbnet: Noise basis learning for image denoising with subspace projection.
\newblock In \emph{Proceedings of the IEEE/CVF conference on computer vision and pattern recognition}, 4896--4906.

\bibitem[{Chi, Jiang, and Mu(2020)}]{chi2020fast}
Chi, L.; Jiang, B.; and Mu, Y. 2020.
\newblock Fast fourier convolution.
\newblock \emph{Advances in Neural Information Processing Systems}, 33: 4479--4488.

\bibitem[{Dai et~al.(2020)Dai, Liu, Li, and Chen}]{dai2020awnet}
Dai, L.; Liu, X.; Li, C.; and Chen, J. 2020.
\newblock Awnet: Attentive wavelet network for image isp.
\newblock In \emph{Computer Vision--ECCV 2020 Workshops: Glasgow, UK, August 23--28, 2020, Proceedings, Part III 16}, 185--201. Springer.

\bibitem[{Hu, Chen, and Allebach(2022)}]{hu2022joint}
Hu, L.; Chen, H.; and Allebach, J.~P. 2022.
\newblock Joint multi-scale tone mapping and denoising for HDR image enhancement.
\newblock In \emph{Proceedings of the IEEE/CVF Winter Conference on Applications of Computer Vision}, 729--738.

\bibitem[{Ignatov et~al.(2021)Ignatov, Chiang, Kuo, Sycheva, and Timofte}]{ignatov2021learned}
Ignatov, A.; Chiang, C.-M.; Kuo, H.-K.; Sycheva, A.; and Timofte, R. 2021.
\newblock Learned smartphone isp on mobile npus with deep learning, mobile ai 2021 challenge: Report.
\newblock In \emph{Proceedings of the IEEE/CVF Conference on Computer Vision and Pattern Recognition}, 2503--2514.

\bibitem[{Ignatov et~al.(2020)Ignatov, Timofte, Zhang, Liu, Wang, Zuo, Zhang, Zhang, Peng, Ren et~al.}]{ignatov2020aim}
Ignatov, A.; Timofte, R.; Zhang, Z.; Liu, M.; Wang, H.; Zuo, W.; Zhang, J.; Zhang, R.; Peng, Z.; Ren, S.; et~al. 2020.
\newblock Aim 2020 challenge on learned image signal processing pipeline.
\newblock In \emph{Computer Vision--ECCV 2020 Workshops: Glasgow, UK, August 23--28, 2020, Proceedings, Part III 16}, 152--170. Springer.

\bibitem[{Ignatov, Van~Gool, and Timofte(2020)}]{ignatov2020replacing}
Ignatov, A.; Van~Gool, L.; and Timofte, R. 2020.
\newblock Replacing mobile camera isp with a single deep learning model.
\newblock In \emph{Proceedings of the IEEE/CVF Conference on Computer Vision and Pattern Recognition Workshops}, 536--537.

\bibitem[{Johnson, Alahi, and Fei-Fei(2016)}]{johnson2016perceptual}
Johnson, J.; Alahi, A.; and Fei-Fei, L. 2016.
\newblock Perceptual losses for real-time style transfer and super-resolution.
\newblock In \emph{Computer Vision--ECCV 2016: 14th European Conference, Amsterdam, The Netherlands, October 11-14, 2016, Proceedings, Part II 14}, 694--711. Springer.

\bibitem[{Li et~al.(2023)Li, Guo, Zhou, Liang, Zhou, Feng, and Loy}]{DBLP:conf/iclr/LiGZLZFL23}
Li, C.; Guo, C.; Zhou, M.; Liang, Z.; Zhou, S.; Feng, R.; and Loy, C.~C. 2023.
\newblock Embedding Fourier for Ultra-High-Definition Low-Light Image Enhancement.
\newblock In \emph{The Eleventh International Conference on Learning Representations, {ICLR} 2023, Kigali, Rwanda, May 1-5, 2023}. OpenReview.net.

\bibitem[{Liang et~al.(2021)Liang, Cai, Cao, and Zhang}]{liang2021cameranet}
Liang, Z.; Cai, J.; Cao, Z.; and Zhang, L. 2021.
\newblock Cameranet: A two-stage framework for effective camera isp learning.
\newblock \emph{IEEE Transactions on Image Processing}, 30: 2248--2262.

\bibitem[{Liu et~al.(2020)Liu, Jia, Liu, and Tian}]{liu2020joint}
Liu, L.; Jia, X.; Liu, J.; and Tian, Q. 2020.
\newblock Joint demosaicing and denoising with self guidance.
\newblock In \emph{Proceedings of the IEEE/CVF Conference on Computer Vision and Pattern Recognition}, 2240--2249.

\bibitem[{Liu et~al.(2018)Liu, Zhang, Zhang, Lin, and Zuo}]{liu2018multi}
Liu, P.; Zhang, H.; Zhang, K.; Lin, L.; and Zuo, W. 2018.
\newblock Multi-level wavelet-CNN for image restoration.
\newblock In \emph{Proceedings of the IEEE conference on computer vision and pattern recognition workshops}, 773--782.

\bibitem[{Ramanath et~al.(2005)Ramanath, Snyder, Yoo, and Drew}]{Ramanath_Snyder_Yoo_Drew_2005}
Ramanath, R.; Snyder, W.; Yoo, Y.; and Drew, M. 2005.
\newblock Color image processing pipeline.
\newblock \emph{IEEE Signal Processing Magazine}, 22(1): 34–43.

\bibitem[{Schwartz, Giryes, and Bronstein(2018)}]{schwartz2018deepisp}
Schwartz, E.; Giryes, R.; and Bronstein, A.~M. 2018.
\newblock Deepisp: Toward learning an end-to-end image processing pipeline.
\newblock \emph{IEEE Transactions on Image Processing}, 28(2): 912--923.

\bibitem[{Sun et~al.(2018)Sun, Yang, Liu, and Kautz}]{sun2018pwc}
Sun, D.; Yang, X.; Liu, M.-Y.; and Kautz, J. 2018.
\newblock Pwc-net: Cnns for optical flow using pyramid, warping, and cost volume.
\newblock In \emph{Proceedings of the IEEE conference on computer vision and pattern recognition}, 8934--8943.

\bibitem[{Suvorov et~al.(2022)Suvorov, Logacheva, Mashikhin, Remizova, Ashukha, Silvestrov, Kong, Goka, Park, and Lempitsky}]{suvorov2022resolution}
Suvorov, R.; Logacheva, E.; Mashikhin, A.; Remizova, A.; Ashukha, A.; Silvestrov, A.; Kong, N.; Goka, H.; Park, K.; and Lempitsky, V. 2022.
\newblock Resolution-robust large mask inpainting with fourier convolutions.
\newblock In \emph{Proceedings of the IEEE/CVF winter conference on applications of computer vision}, 2149--2159.

\bibitem[{Wang et~al.(2018)Wang, Yu, Dong, and Loy}]{wang2018recovering}
Wang, X.; Yu, K.; Dong, C.; and Loy, C.~C. 2018.
\newblock Recovering realistic texture in image super-resolution by deep spatial feature transform.
\newblock In \emph{Proceedings of the IEEE conference on computer vision and pattern recognition}, 606--615.

\bibitem[{Wang, Simoncelli, and Bovik(2003)}]{wang2003multiscale}
Wang, Z.; Simoncelli, E.~P.; and Bovik, A.~C. 2003.
\newblock Multiscale structural similarity for image quality assessment.
\newblock In \emph{The Thrity-Seventh Asilomar Conference on Signals, Systems \& Computers, 2003}, volume~2, 1398--1402. Ieee.

\bibitem[{Yang et~al.(2023)Yang, Cheng, Zhao, Yan, Zhang, and Li}]{yang2023learning}
Yang, K.-F.; Cheng, C.; Zhao, S.-X.; Yan, H.-M.; Zhang, X.-S.; and Li, Y.-J. 2023.
\newblock Learning to adapt to light.
\newblock \emph{International Journal of Computer Vision}, 131(4): 1022--1041.

\bibitem[{Zhang et~al.(2018)Zhang, Isola, Efros, Shechtman, and Wang}]{zhang2018unreasonable}
Zhang, R.; Isola, P.; Efros, A.~A.; Shechtman, E.; and Wang, O. 2018.
\newblock The unreasonable effectiveness of deep features as a perceptual metric.
\newblock In \emph{Proceedings of the IEEE conference on computer vision and pattern recognition}, 586--595.

\bibitem[{Zhang et~al.(2021)Zhang, Wang, Liu, Wang, Zhang, and Zuo}]{zhang2021learning}
Zhang, Z.; Wang, H.; Liu, M.; Wang, R.; Zhang, J.; and Zuo, W. 2021.
\newblock Learning raw-to-srgb mappings with inaccurately aligned supervision.
\newblock In \emph{Proceedings of the IEEE/CVF International Conference on Computer Vision}, 4348--4358.

\bibitem[{Zhou and Glotzbach(2007)}]{zhou2007image}
Zhou, J.; and Glotzbach, J. 2007.
\newblock Image pipeline tuning for digital cameras.
\newblock In \emph{2007 IEEE International Symposium on Consumer Electronics}, 1--4. IEEE.

\bibitem[{Zhou et~al.(2023)Zhou, Huang, Guo, and Li}]{zhou2023fourmer}
Zhou, M.; Huang, J.; Guo, C.-L.; and Li, C. 2023.
\newblock Fourmer: An Efficient Global Modeling Paradigm for Image Restoration.
\newblock In \emph{International Conference on Machine Learning}, 42589--42601. PMLR.

\bibitem[{Zhou et~al.(2022)Zhou, Huang, Yan, Yu, Fu, Liu, Wei, and Zhao}]{zhou2022spatial}
Zhou, M.; Huang, J.; Yan, K.; Yu, H.; Fu, X.; Liu, A.; Wei, X.; and Zhao, F. 2022.
\newblock Spatial-frequency domain information integration for pan-sharpening.
\newblock In \emph{European Conference on Computer Vision}, 274--291. Springer.

\end{thebibliography}

\end{document}